\pdfoutput=1

\documentclass[11pt]{article}

\usepackage{emnlp2022}

\usepackage{times}
\usepackage{latexsym}

\usepackage[T1]{fontenc}

\usepackage[utf8]{inputenc}

\usepackage{microtype}

\usepackage{inconsolata}

\usepackage{tikz-dependency}
\usepackage{verbatim}
\usepackage{amsmath}
\usepackage{epsfig}
\usepackage{multirow}
\usepackage{linguex}
\usepackage{graphicx}
\usepackage{subfig}
\usepackage{gb4e}
\usepackage{fullpage}
\usepackage{wrapfig}
\usepackage{cases}
\usepackage{float}
\usepackage{color}
\usepackage{booktabs}
\usepackage{nameref}
\usepackage[titletoc]{appendix}
\usepackage{amsmath}
\usepackage{lipsum}  

\title{Dual Mechanism Priming Effects in Hindi Word Order}

\author{Sidharth Ranjan \\
  IIT Delhi\\
  \texttt{sidharth.ranjan03@gmail.com} \\\And
  Marten van Schijndel \\
  Cornell University \\
  \texttt{mv443@cornell.edu} \\ \AND
  Sumeet Agarwal \\
  IIT Delhi \\
  \texttt{sumeet@iitd.ac.in} \\\And
  Rajakrishnan Rajkumar \\
  IISER Bhopal \\
  \texttt{rajak@iiserb.ac.in} \\}

\begin{document}
\maketitle

\begin{abstract}





Word order choices during sentence production can be primed by preceding sentences. In this work, we test the \textsc{dual mechanism} hypothesis that priming is driven by multiple different sources. Using a Hindi corpus of text productions, we model lexical priming with an n-gram cache model and we capture more abstract syntactic priming with an adaptive neural language model. We permute the preverbal constituents of corpus sentences, and then use a logistic regression model to predict which sentences actually occurred in the corpus against artificially generated meaning-equivalent variants. Our results indicate that lexical priming and lexically-independent syntactic priming affect complementary sets of verb classes. By showing that different priming influences are separable from one another, our results support the hypothesis that multiple different cognitive mechanisms underlie priming.

\end{abstract}

\section{Introduction}\label{sect:intro}

\citet{gries2005} defines syntactic priming as the tendency of speakers ``to repeat syntactic structures they have just encountered (produced or comprehended) before''. Starting with~\citet{bock1986syntactic}, a long line of experimental and corpus-based work has provided evidence for this phenomenon in the context of language production~\citep[see][for a through review]{reitter2011}. More recently, comprehension studies have also attested priming effects in a wide variety of languages~\citep{arai2007priming,tooley2010syntactic}, where prior experience of a syntactic structure alleviates the comprehension difficulty associated with subsequent similar syntactic structures during reading. The experimental record also demonstrates that lexical repetition affects syntactic priming~\citep[][and references therein]{reitter2011}. According to the \textsc {dual mechanism account} proposed by~\citet{tooley2010syntactic}, lexically independent syntactic priming effects are caused by an implicit learning mechanism~\citep{BockGriff2000,chang2006}, whereas lexically dependent priming effects are caused by a more short-term mechanism, such as residual activation~\citep{Pickering1998}.

In the present work, we test this hypothesis of a dual mechanism of priming by analyzing whether different kinds of intersentential priming can account for the word order of different constructions in Hindi. Our main contribution is that we deploy precisely defined quantitative cognitive factors in our statistical models along with minimally paired alternative productions, whereas most previous experimental and corpus studies on priming only employ one or the other.

Hindi has a flexible word order, though SOV is the canonical order \citep{kachru2006hindi}. To investigate constituent ordering preferences, we generate meaning-equivalent grammatical variants of Hindi sentences by linearizing preverbal constituents of projective dependency trees of the Hindi-Urdu Treebank corpus \citep[HUTB;][]{Bhatt2009} of written text. 
We validated the assumptions underlying this method using crowd-sourced human judgments and compared the performance of our machine learning model with the choices made by human subjects.
Pioneering studies of Hindi word order have demonstrated a wide variety of factors that influence order preferences, such as information status~\citep{buttking1996,kidwai2000}, prosody~\citep{focusHindi08}, and semantics~\citep{PereraSrivastava2016,mohanan1994wordorder}. We incorporated measures of these baseline influences into a logistic regression model to distinguish the original reference sentences from our generated variants.
We model lexical priming with an n-gram cache model and we capture more abstract syntactic priming with an adaptive neural language model. \citet{gries2005} showed that syntactic priming effects are strongly contingent on verb class. To this end, we analyze model behavior on sentences involving the following verb classes: Levin's~\citeyearpar{levin1993english} syntactic-semantic verb classes, verbs involved in double object constructions, and conjunct verbs involving noun-verb complex predicates. 
To foreshadow our results, information-theoretic surprisal computed using our two different models predicts word order in complementary linguistic contexts over the baseline predictors. Moreover, for the task of choosing reference vs variant sentences, the model's predicted choices matched the agreement between human subjects for all of Levin's verb classes. By showing that different priming influences are separable from one another, our results support the dual mechanism hypothesis that multiple different cognitive mechanisms underlie priming.

\section{Data}\label{sect:method}

Our data set consists of 1996 reference sentences containing well-defined subject and object constituents corresponding to the projective dependency trees in HUTB corpus~\citep{Bhatt2009}. The sentences in the HUTB corpus belong to the newswire domain and contain written text in a naturally occurring context \emph{i.e.,} every sentence in the news article was situated in the context of preceding sentences. For each reference sentence in our data set, we created counterfactual grammatical variants expressing the same truth-conditional meaning\footnote{A limitation of this definition: It does not capture the fact that, in contrast to marked orders, which necessitate context for a full interpretation, SOV canonical orders are neutral with respect to the preceding discourse~\citep{gambhirphd}.} 
by permuting the preverbal constituents whose heads were linked to the root node in the dependency tree.\footnote{Appendix~\ref{appendix:A} explains our variant generation procedure in more detail.} 
Inspired by grammar rules proposed in the NLG literature~\citep{llc:raja:mwhite:2014}, ungrammatical variants were automatically filtered out by detecting dependency relation sequences not attested in the original HUTB corpus. After filtering, we had 72833 variant sentences for our classification task. 

\section{Classification Task}\label{subsect:model}


In order to mitigate the data imbalance between the two groups (1996 references vs. 72833 variants), we follow \citet{Joachims:2002} by formulating our task as a pair-wise ranking problem. 

\vspace{-1.5em}
\begin{align}\label{eq:nor}
 w~\cdot~\phi(reference) > w~\cdot~\phi(variant)\\
 \label{eq:joc}
 w~\cdot~(\phi(reference)~-~\phi(variant)) > 0  
\end{align}

The goal of the basic binary classifier model is shown in Equation \ref{eq:nor}, where the model learns a feature weight ($w$) such that the dot product of the variant feature vector ($\phi(variant)$) with $w$ is less than the dot product of $w$ with the reference feature vector ($\phi(reference)$). The same goal can be written as Equation \ref{eq:joc} which ensures that $w$'s dot product with the difference between the feature vectors is positive. This transformation alleviates issues from having dramatically unbalanced class distributions.


We first arranged the references and variants into ordered pairs (e.g., a reference with two variants would be paired as ($reference$, $variant_1$) and ($variant_2$, $reference$)), and then subtracted the feature vectors of the first member of the pair from the feature vectors of its second member. We then assigned binary labels to each pair, with \textit{reference-variant} pairs coded as ``1'', and \textit{variant-reference} pairs coded as ``0'', thus re-balancing our previously severely imbalanced classification task. Additionally, the feature values of sentences with varying lengths get centered using this technique. Refer to \citet{cog:raja} and \citet{cog:sid} for a more detailed illustration.

Using features extracted from the transformed dataset, we trained a logistic regression model to predict each reference sentence (see Equation \ref{eq:regr}). All the experiments were done with the Generalized Linear Model (GLM) package in $R$. Here \textit{choice} is encoded by the binary dependent variable as discussed above ($1$: reference preference and $0$: variant preference).

\vspace{-0.5em}
\begin{small}
\begin{equation}\label{eq:regr}
choice \sim  \begin{cases}
 & \text{$\delta$ dependency length +} \\ 
 & \text{$\delta$ trigram surp + $\delta$ pcfg surp +} \\ 
 & \text{$\delta$ IS score + $\delta$ lexical repetition surp +}\\ 
 & \text{$\delta$ lstm surp + $\delta$ adaptive lstm surp}
\end{cases}
\end{equation}
\end{small}

\subsection{Cognitive Theories and Measures}\label{sect:cog-meas}

\subsubsection{Surprisal Theory}

According to the Surprisal Theory~\citep{hale2001,levy2008}, comprehenders build probabilistic interpretations of phrases based on patterns they have already seen in sentence structures. Mathematically, the \textit{surprisal} of the ${k}^{th}$ word, $w_k$, is defined as the negative log probability of $w_k$ given the preceding context: 

\begin{equation}{\label{eq1}}
S_{k}= -\log P(w_{k}|w_{1...k-1})
\end{equation}


These probabilities, which indicate the information load (or predictability) of $w_k$, can be calculated over word sequences or syntactic configurations. The theory is supported by a large number of empirical evidences from behavioural as well as broad-coverage corpus data comprising both comprehension~\cite{DembergKeller2008,boston2008parsing,Roark2009,cog:sid,staub2015,agrawal2017expectation} and production modalities~\cite{demberg-sayeed-2012,dammalapati2021expectation,dammalapati-etal-2019-expectation,ranjan-etal-2019-surprisal,scil:sid,jain-etal-2018-uniform}.

Using the above surprisal framework, we estimate various types of surprisal scores for each test sentence in our dataset as described below serving as independent variables in our experiment. The word-level surprisal of all the words in each sentence were summed to obtain sentence-level surprisal measures. 

\begin{enumerate}

  \item  {\bf Trigram surprisal}: We calculated the local predictability of each word in a sentence using a 3-gram language model (LM) trained on 1 million sentences of mixed genre from the EMILLE Hindi corpus \citep{emille2002} using the SRILM toolbox \citep{SRILM-ICSLP:2002} with Good-Turing discounting. 

  \item {\bf PCFG surprisal}: We estimated the syntactic probability of each word in the sentence using the Berkeley latent-variable PCFG parser\footnote{5-fold CV parser training and testing F1-score metrics were 90.82\% and 84.95\%, respectively.}~\citep{bkp2006}. We created 12000 phrase structure trees by converting HUTB dependency trees into constituency trees using the approach described in \citet{Yadav2017KeepingIS}. Subsequently, we used them to train the Berkeley PCFG parser. Sentence level log-likelihood of each test sentence was estimated by training a PCFG language model on four folds of the phrase structure trees and then testing on a fifth held-out fold.

 \item \textbf{Lexical repetition surprisal}: Following the method proposed by \citet{kuhn1990cache}, we estimated cache-based surprisal of each word in a sentence using SRILM toolbox by interpolating a 3-gram LM with a unigram cache LM based on the history of words ($|H|=100$) involving the preceding sentence with a default interpolation weight parameter ($\mu=0.05$; see Equations \ref{cache1} and \ref{cache2}). The basic idea is to keep track of word tokens that appeared recently and then amplify their likelihood of occurrence in the trigram word sequence. In other words, the following sentences are more likely to use words again that have recently appeared in the text~\citep{kuhn1990cache,Clarkson97languagemodel}. This way, we account for the lexical priming effect in sentence processing.


\begin{small}
\begin{equation}{\label{cache1}}
\begin{aligned}
P(w_{k}|w_{1..k-1}) = \mu ~P_{cache}(w_{k}|w_{1..k-1})\\ + (1-\mu)~{P_{trigram}(w_{k}|w_{k-2},w_{k-1})}
\end{aligned}
\end{equation}
\begin{equation}{\label{cache2}}
\begin{aligned}
P_{cache}(w_{k}|w_{1..k-1}) = \frac{count_H(w_{k})}{|H|}
\end{aligned}
\end{equation}
\end{small}

 \item \textbf{LSTM surprisal}: The probabilities of each word in the sentence were estimated according to the entire sentence prefix using a long short-term memory language model~\citep[LSTM;][]{hochreiter1997long} trained on 1 million sentences of the EMILLE Hindi corpus. We used the implementation provided in the neural complexity toolkit\footnote{\url{https://github.com/vansky/neural-complexity}}~\citep{van2018neural} with default hyper-parameter settings to estimate surprisal using an unbounded neural context.
 
 \item  \textbf{Adaptive LSTM surprisal}: Following the method proposed by \citet{van2018neural}, we calculated the discourse-enhanced surprisal of each word in the sentence. The cited authors presented a simple way to continuously adapt a neural LM, and found that adaptive surprisal considerably outperforms non-adaptive surprisal at predicting human reading times. They use a pre-trained LSTM LM and, after estimating surprisal for a test sentence, change the LM's parameters based on the sentence's cross-entropy loss. 
 After that, the revised LM weights are used to predict the next test sentence. In our work, we estimated the surprisal scores for each test sentence using neural complexity toolkit by adapting our base (non-adaptive) LSTM LM to one preceding context sentence.

\end{enumerate}

\subsubsection{Dependency Locality Theory}

Shorter dependencies are typically simpler to process than longer ones, according to the Dependency Locality Theory \citep{gibson00}, which has been demonstrated to be effective at predicting the comprehension difficulty of a sequence \citep[][cf.\ \protect\citeauthor{DembergKeller2008}, \citeyear{DembergKeller2008}]{Temperley2007,futrell2015,Liu2017}. Following the work by \citet{Temperley08} and \citet{cog:raja}, we calculated sentence-level dependency length by summing the head-dependent distances (measured as the number of intervening words) in the dependency trees of reference and variant sentences.

\subsubsection{Information Status}

Languages generally prefer to mention \textit{given} referents, from earlier in the discourse, before introducing \textit{new} ones~\citep{Clark-Haviland77,chafe1976givenness,kaiser2004role}. We assigned a \textit{Given} tag to the subject and object constituents in a sentence if any content word within them was mentioned in the preceding sentence or if the head of the phrase was a pronoun. All other phrases were tagged as \textit{New}. For each sentence, IS score was computed as follows: a)~Given-New order = +1 b)~New-Given order = -1 c)~Given-Given and New-New = 0. For illustration, see Appendix~\ref{appendix:B}, which shows how givenness would be coded after a context sentence.

\section{Experiments and Results}\label{sect:results}

We tested the hypothesis that surprisal enhanced with inter-sentential discourse information (adaptive LSTM surprisal) predicts constituent ordering in Hindi over other baseline cognitive controls, including information status, dependency length, lexical repetition, and non-adaptive surprisal. For our adaptation experiments, we used an adaptive learning rate of 2 as it minimized the perplexity of the validation data set (see Table~\ref{tab:lr-rate} in Appendix~\ref{appendix:lr-adapt}). The Pearson's correlation coefficients between different predictors are displayed in Figure \ref{fig:corr-plot} in Appendix~\ref{appendix:pearson}. The adaptive LSTM surprisal has a high correlation with all other surprisal features and a low correlation with dependency length and information status score. On specific verbs of interest, we report the results of the regression and prediction experiments (using 10-fold cross-validation, i.e., a model trained on 9 folds was used to generate predictions on the remaining fold). A prediction experiment using feature ablation helped ascertain the impact of syntactic priming independent of lexical repetition effects. We conducted a fine-grained verb-specific analysis of priming patterns on conjunct verbs and Levin's syntactic-semantic classes, followed by a targeted human evaluation of Levin's verb classes.





\subsection{Verb-Specific Priming}

Individual verb biases are well known to influence structural choices during language production~\citep{ferreira2013verb,malathiverb2017,YiKoenigRoland2019} and priming effects are also contingent on specific verbs~\citep{gries2005}. Therefore, we grouped Hindi verbs based on \citeauthor{levin1993english}'s syntactico-semantic classes using the heuristics proposed by \citet{begum2017development}. Then we analyzed the efficacy of adaptive surprisal at classifying reference and variant instances of Levin's verb classes (still training the classifier on the full training partition for each fold). Our results (Table \ref{tab:verb-arg:pred}, top block) indicate that the \textsc{give} verb class was susceptible to priming, with adaptive surprisal producing a significant improvement of 0.12\% in classification accuracy (p = 0.01 using McNemar’s two-tailed test) over the baseline model. The regression coefficients pertaining to \citeauthor{levin1993english}'s \textsc{give} verb classes are presented in Table \ref{tab:regr-result-give} in Appendix~\ref{appendix:verb-reg}. Other Levin verb frames did not show syntactic priming. 


\begin{table}[t]
\begin{small}
\begin{tabular}{ll|cc}
\toprule
  \textbf{Type} & \textbf{Freq} & \textbf{Baseline} & \textbf{Baseline +} \\
                    & (\%) &   & \textbf{  Adaptive LSTM} \\
                      \midrule
{\it Verb Class} & & & \\
  \textsc{do} & 48.68 & 96.82 & 96.82 \\
\textsc{\textbf{give}} & 19.35 & 93.86 & \textbf{93.98} \\
\textsc{social} & 8.00 & 92.90 & 92.95 \\
\textsc{communicate} & 6.25 & 93.94 & 93.98 \\
\textsc{lodge} & 4.04 & 94.29 & 94.22 \\
\textsc{motion} & 3.87 & 90.87 & 90.76 \\
\textsc{put} & 2.97 & 95.28 & 95.28 \\
\textsc{destroy} & 2.42 & 95.58 & 95.63 \\
\textsc{perception} & 0.73 & 87.48 & 87.10 \\
\textsc{others} & 3.69 & 90.63 & 90.22\\\midrule
{\it Alternations} & & & \\
{S-DO} & 71.89 & 95.35 & 95.33 \\
\textbf{S-IO-DO} & 12.74 & 93.39 & \textbf{93.50} \\
{S-IO} & 15.37 & 94.98 & 95.04\\\bottomrule

\end{tabular}
\end{small}
\caption{Prediction performance of verb-specific and subject-objects alternations (72833 points); Baseline denotes \textit{base1} shown in Table \ref{tab:lex-adapt-pred-acc}; bold denotes McNemar's two-tailed significance compared to baseline model in the same row)}
\label{tab:verb-arg:pred}
\end{table}

Our results align with previous work in the priming literature that shows \textsc{give} to be especially susceptible to priming, thus providing cross-linguistic support to verb-based priming effects~\citep{Pickering1998,gries2005,bock1986syntactic}. The \textsc{give} verb class in our data set includes different verbs that are semantically similar to \textit{give} in English, such as 
\textit{de, saup, bhej, maang, dila, lautaa, vasul, thama, vaapas}. We found that all these verbs strongly exhibited double object constructions~\citep{begum2017development} and their arguments are often case marked (see Table \ref{tab:verb:case} in Appendix \ref{appendix:verbcd} for more details). 

\subsection{Double Object construction}

Previous studies on dative alternations in psycholinguistics have shown that the propensity of speakers to produce such constructions increases with their recent mention~\citep{bock1986syntactic,kaschak2006recent}. The same factors also influence their predictability in reading comprehension~\citep{van2018neural,tooley2010syntactic,tooley2014parity}. To test whether such effects determine word-ordering decisions in Hindi, we isolated double object constructions from our dataset such that the main verb compulsorily has two objects \emph{viz.,} direct and indirect objects in the sentence. Table~\ref{tab:regr-result-k1k2k4} shows that all predictors (including adaptive and lexical repetition surprisal) are significant predictors of syntactic choice.


Then we analyzed the efficacy of adaptive surprisal at classifying reference and variant instances of double object constructions (still training the classifier on the full training partition for each fold). We also conducted a comparison of our results with single-object constructions. Our results (Table \ref{tab:verb-arg:pred}, bottom block) reveal that syntactic priming effects are present over and above lexical repetition effects. Syntactic priming is more influential in double object constructions (S-IO-DO) than in single object constructions (S-IO or S-DO), as attested by a significant improvement of 0.1\% in classification accuracy (p = 0.04 using McNemar’s two-tailed test). Double object constructions are also highly case marked (see Table \ref{tab:arg:case} in Appendix~\ref{appendix:k1k2k4cd}) and 57.82\% of these items contain verbs that belong to \textsc{give} class (see Table~\ref{tab:verb:k1k2k4} in Appendix~\ref{appendix:verbclass} for more details). In the discussion section we present a more nuanced discussion on the effects of case-markers and a verb's combinatorial properties on priming.

\begin{table}[t]
\centering
\scalebox{1.1}{
  \begin{small}
\begin{tabular}{lccc}
\toprule
Predictor & $\hat\beta$ & $\hat\sigma$ & t\tabularnewline
\midrule
intercept   & \textbf{1.50} & 0.003 & 506.77 \\
trigram surprisal  & \textbf{-0.14} & 0.017 & -8.30 \\
dependency length  & \textbf{0.02} & 0.003 & 6.20 \\
pcfg surprisal  & \textbf{-0.11} & 0.005 & -20.8 \\
IS score  & \textbf{0.02} & 0.003 & 5.43 \\
lex-rept surprisal & \textbf{0.06} & 0.016 & 4.07 \\
lstm surprisal  & \textbf{0.31} & 0.081 & 3.81 \\
adaptive lstm surprisal  & \textbf{-0.59} & 0.081 & -7.23\\
\bottomrule 
\end{tabular}
  \end{small}
  }
\caption{Regression model on double object construction \textsc{S-IO-DO} data set (9278 data points; all significant predictors denoted by $|$t$|$\textgreater{}2)}
\label{tab:regr-result-k1k2k4}
\end{table}

In summary, our analyses suggest that different verbs display varying strengths of priming effects, corroborating previous findings in the literature~\citep{gries2005}. Ditransitive constructions (denoted by S-IO-DO ordering) prime more strongly than other orderings, where verbs from the \textsc{give} class strongly prefer canonical argument ordering\footnote{For example, out of 284 instances, 89.79\% of the lemma `de' (\textsc{give} class) occurs with canonical argument ordering in our test data set.} while determining Hindi syntactic choices.

\subsection{Example Analysis: Success of Adaptive LSTM Surprisal}\label{sect:qual-lstmadapt}




We now discuss the example below to illustrate discourse-based syntactic priming effects (estimated via adaptive surprisal) in determining the preferred syntactic choice among referent-variant pairs (\ref{ex:ref}, \ref{ex:var}). 
\newpage

\begin{small}
\begin{exe}

  \ex \label{ex:hindi-verb-arg-prev}
  {\label{ex:h1p} {\bf Context Sentence}
    \gll {collingwood 8} {aur} {jones 0} {aur} {blackville 10-par} {hi} {harbhajan-ki} {firki-ka} {sikaar \textbf{ban gaye}}\\
         {collingwood 8} {and} {jones 0} {and} {blackville 10-\textsc{psp}} {\textsc{emph}} {harbhajan-\textsc{gen}} {spin-\textsc{gen}} {victim \textbf{become}-\textsc{pst}}\\
     \glt \begin{small}{\it Collingwood became the victim of Harbhajan's spin on 8 and Jones on 0 and Blackville on just 10.}\end{small}}

\end{exe}

\begin{exe}

  \ex \label{ex:hindi-verb-arg}
  \begin{xlist}
  \ex[]{\label{ex:ref}
    \gll {\it plunket} {\underline{14-par}} {pathan-ki} {gend-par} {Gambhir-ko} {kaetch} {\bf de baethe} \textbf{(Reference)}\\
         {plunket} {14-\textsc{psp}} {pathan-\textsc{gen}} {ball-\textsc{psp}} {gambhir-\textsc{gen}} {catch} {give\textsc{.pst.sg}}\\
     \glt \textit{Plunket ended up giving a catch to Gambhir on 14 off Pathan's bowling.}\\}
  \ex[] {\label{ex:var} {\underline{14-par}} {\it plunket} {pathan-ki} {gend-par} {gambhir-ko} {kaetch} {\bf de baethe} \textbf{(Variant)}}
  \end{xlist}

\end{exe}
\end{small}

The LSTM LM when adapted to the previous sentence (\ref{ex:hindi-verb-arg-prev}) and tested on referent-variant pairs (\ref{ex:hindi-verb-arg}) assigns a lower surprisal to the reference sentence (\ref{ex:ref}) than its competing variant (\ref{ex:var}). It is conceivable that the adaptive LSTM suprisal learns syntactic patterns in the context sentence and prefers the reference sentence (over the variant) owing to the 
similarities between the reference and context sentences. Every other predictor aside from adaptive LSTM surprisal fails to predict the corpus reference sentence over the paired variant, in spite of the fact that the reference sentence has canonical ordering and the alternative variant has non-canonical ordering. This could be attributed to multiple factors. For example, dependency length would prefer the variant since the long-short sequence ($14~par$-$plunket$) in the variant minimizes its dependency length unlike the short-long sequence ($plunket$-$14~par$) in the reference sentence. Similarly, the intra-sentential surprisal models make the wrong choice while processing the sentences because they possibly get locally garden pathed due to the two consecutive proper nouns (NPs) \emph{viz.,} $plunket$ and $pathan$ (referring to 2 distinct individuals in the real world as opposed to \textit{plunket pathan} referring to a single individual). Table \ref{tab:sent-level-scores} and Figure \ref{fig:k1k2k4-profile} in Appendix~\ref{appendix:k1k2k4} present the sentence-level predictor values of reference-variant pairs (Example~\ref{ex:hindi-verb-arg}) and their information profiles respectively illustrating these patterns. 





\subsection{Conjunct Verb Construction}


\begin{table}[t]
\centering
\begin{small}
\scalebox{1.1}{
\begin{tabular}{lccc}
\toprule
Predictor & $\hat\beta$ & $\hat\sigma$ & t\tabularnewline
\midrule 
intercept  & 1.50 & 0.001 & 1379.73 \\
trigram surprisal  & -0.09 & 0.005 & -15.27 \\
dependency length  & 0.01 & 0.001 & 7.82 \\
pcfg surprisal  & -0.07 & 0.002 & -35.55 \\
IS score  & 0.02 & 0.001 & 13.70 \\
lex-rept surprisal  & -0.02 & 0.005 & -2.98 \\
lstm surprisal  & -0.14 & 0.016 & -8.60 \\
adaptive lstm surprisal  & -0.12 & 0.016 & -7.40\\
\bottomrule 
\end{tabular}}
\caption{Regression model on conjunct verb data set ($N=51617$; all significant predictors denoted by $|$t$|$\textgreater{}2)}
\label{tab:regr-results-cv}
\end{small}
\end{table}

In this section, we go beyond \citeauthor{levin1993english}'s verb class and study the effects of priming on sentences containing conjunct verbs. Hindi conjunct verbs are \textsc{noun-verb} complex predicates (CP) in which a highly predictable verb follows a nominal leading to a non-compositional meaning~\citep{butt1995structure,mohanan1994argument,husain2014expectations}. For example, the complex predicates, such as 
\textit{khyaal rakhna} (`care keep/put'; `to take care of') with non-compositional meaning are associated with conjunct verb construction in our dataset (marked with the \textsc{pof} dependency relation label in the HUTB corpus) unlike the predicate 
\textit{guitar rakhna} (`guitar keep/put'; `to put down or keep a guitar') that has compositional meaning.  

In particular, we examined the impact of adaptive LSTM surprisal in predicting corpus reference sentences amidst the variants on the subset of the data consisting of conjunct verbs. Prior work in sentence comprehension has investigated the effects of expectation and locality in Hindi conjunct verb constructions~\citep{husain2014expectations,cog:sid}. The conjunct verb subset in our dataset contains 40.68\% of reference sentences out of 1996, leading to 51,617 data points (referent-variant pairs) for our classification task.

Our regression results (Table \ref{tab:regr-results-cv}) demonstrate that all the measures considered in our work are significant predictors of syntactic choice in Hindi. The negative regression coefficient of adaptive LSTM surprisal indicates that noun-verb predicate structures are more common in the context of similarly occurring noun-verb predicate structures, thus providing preliminary indication of potential priming effects. Further corpus analysis revealed that 35\% of conjunct verb marked context sentences preceded reference sentences with conjunct verb phrases in our dataset. Adding adaptive LSTM surprisal into the regression model containing all other predictors significantly improved the fit ($\chi^2$ = 187.27; p $<$ 0.001).

We now examine the relative performance of adaptive LSTM surprisal on conjunct verb constructions above and beyond every other feature in the classification model. We also conduct a feature ablation study to ascertain the impact of syntactic priming (adaptive LSTM surprisal) independent of lexical priming (lexical repetition surprisal) in determining syntactic choices in Hindi. We used the model trained over the entire training partition for each fold from the full dataset and then tested only on the conjunct-verb test partition. 
We found that even for conjunct verb constructions (rightmost column of Table \ref{tab:lex-adapt-pred-acc} in Appendix \ref{sect:full-data}), adaptive LSTM surprisal induced a significant increase of 0.04\% in prediction accuracy (p = 0.04 using McNemar's two-tailed test) over a baseline comprised of all predictors but lexical repetition surprisal. Adaptive LSTM surprisal ceased to be a general predictor when lexical repetition surprisal was incorporated into the classification model. This result provides an evidence for a generalized \textit{lexical boost effect} in Hindi, which operates over verb classes (conjunct verbs here) and not simply string-identical verbs, validating similar findings in English~\citep{Snider09}. 

Additionally, Table \ref{tab:lex-adapt-pred-acc} in Appendix \ref{sect:full-data} also presents the results of our classification experiment on the full dataset (72833 points). The findings discussed above for conjunct verb construction extend to full data as well. 
Besides, the feature ablation experiments on both full dataset and conjunct verb subset also suggest that when lexical repetition is taken into account there is weak tendency for the individual to repeat their own syntactic construction from preceding contextual sentence except for certain constructions as discussed in the preceding sections. Interestingly, similar findings have been reported for English dialogue corpora as well~\citep{healey2014divergence,green2021global}. Future work needs to perform principled investigation on Hindi spoken data to understand the divergence and commonalities among written and verbal communication, and to make more substantial claims about priming in language production.

\subsection{Example Analysis: Success of Lexical Repetition Surprisal}\label{sect:qual-lexrept}

This section discusses the following example where lexical repetition surprisal estimated using {\it n}-gram cache LM is the only predictor that makes the right choice by choosing the reference sentence \ref{ex:ref-lex-rept} and every other measures predict the alternative variant \ref{ex:var-lex-rept} as their preferred syntactic choice. 

\begin{small}
\begin{exe}

  \ex \label{ex:hindi-lex-rept-prev}
  {\label{ex:h1p-lex-rept-prev} {\bf Context Sentence}
    \gll {\underline{jailon}-ki} {jo} {haalat} {hai} {usme} {kisi} {kaedi-ka} {paagal} {ho jana} {maamuli baat hai}\\
         {prisons-\textsc{gen}} {such} {condition} {be} {in that} {any} {prisoner-\textsc{gen}} {insane} {become-\textsc{fut}} {minor thing}\\
     \glt \begin{small}{\it Such are the conditions of prisons, it is a minor thing for any prisoner to go insane.}\end{small}}

\end{exe}

\begin{exe}

  \ex \label{ex:hindi-lex-rept}
  \begin{xlist}
  \ex[]{\label{ex:ref-lex-rept}
    \gll {varshon-tak} {mukadamen-ka} {intejaar} {\underline{jailon}-mein} {sadate} {een} {vichaaraadheen} {kaidiyon-ko} {avasaad-mein} {jaane-ko} {vivash} {kar deti hai} \textbf{(Reference)}\\
         {for years} {trial-\textsc{gen}} {waiting} {prisons-\textsc{loc}} {rotting} {these} {under-trial} {prisoners-\textsc{acc}} {depression-\textsc{loc}} {go-\textsc{inf}} {compel} {do\textsc{.prs.sg}}\\
     \glt \textit{Waiting for trial for years compels these undertrial prisoners rotting in jails to go into depression..}\\}
  \ex[] {\label{ex:var-lex-rept} {\underline{jailon}-mein} {sadate} {een} {vichaaraadheen} {kaidiyon-ko} {avasaad-mein} {jaane-ko} {vivash} {varshon-tak} {mukadamen-ka} {intejaar} {kar deti hai} \textbf{(Variant)}}
  \end{xlist}

\end{exe}
\end{small}

Table \ref{tab:sent-level-scores} in Appendix \ref{appendix:k1k2k4} presents the sentence-level predictor values for referent-variant pairs (Example \ref{ex:hindi-lex-rept}). For both sentences, the trigram cache LM assigns a high probability to the word `jailon' (\textit{prisons)} as the word is mentioned\footnote{In contrast, the examples discussed in Section~\ref{sect:qual-lstmadapt} denoting syntactic priming do not have content word repetition across sentences.} in the preceding context sentence (Example~\ref{ex:hindi-lex-rept-prev}). However, at the sentence level, the cache LM allocates low surprisal score to the reference sentence (\ref{ex:ref-lex-rept}), thus predicting it to be a best choice than the variant sentence (\ref{ex:var-lex-rept}). Altogether, this analysis indicates that lexical repetition surprisal accounts for the word's preference to be in a syntactic configuration where the sequence is more probable, favoring the corpus reference sentence. We also argue that the long subject phrase (\textit{varshon~tak~mukadamen~ka~intejaar}) in the reference sentence is hard to interpret in isolation (\emph{i.e.,} in the absence of the previous context sentence~\ref{ex:hindi-lex-rept-prev}), potentially affecting the intra-sentential surprisal estimation that does not factor in the context information from the preceding sentence. Moreover, due to its long-short constituent and \textsc{new-given} orderings, additional factors like dependency length and IS score do not favor the reference sentence too.

\begin{table}[t]
    \centering
    \scalebox{0.7}{
    \begin{tabular}{l|ccc}
    Levin's verb & Agreement (\%) & Model (\%) & Model (\%) \\
    Type (item count) & human:corpus & corpus & human \\\hline
    \textsc{do} (32) & 84.38 & 65.63 & 68.75\\
    \textsc{social} (30) & 86.67 & 70 & 76.67\\\hline
    \textsc{give} (46) & 86.96 & 67.39 & 67.39\\
    \textsc{communication} (26) & 100 & 92.31 & 92.31\\
    \textsc{motion} (9) & 77.78 & 66.67 & 66.67\\
    \textsc{put} (8) & 100 & 75 & 75\\\hline
    \textsc{lodge} (8) & 100 & 100 & 100\\
    \textsc{perception} (4) & 100 & 100 & 100\\
    \textsc{destroy} (2) & 100 & 100 & 100\\
    \textsc{others} (2) & 100 & 100 & 100\\\hline
    Total (167) & 89.92 & 74.85 & 76.65\\
    \end{tabular}}
    \caption{Targeted human evaluation --- \textbf{Agreement human/corpus}: Percentages of times human judgement matches with corpus reference choice; \textbf{Model corpus}: Percentages of corpus choice correctly predicted by the classifier containing all the predictors; \textbf{Model human:} Percentages of human label correctly predicted by the classifier containing all the predictors}
    \label{tab:human-eval}
\end{table}

\subsection{Targeted Human Evaluation}\label{subsec:human-eval}


We conducted a targeted human evaluation to validate our order-permutation analysis and to compare the choices made by the machine learning model with those of native speakers of Hindi. To this end, we designed a forced-choice task and collected sentence judgments from 12 Hindi native speakers for 167 randomly selected reference-variant pairs in our data set. Participants were first shown the context sentence, and were then asked to judge the most likely following sentence amongst the reference-variant pair. Each sentence was assigned a human label of ``1'' if more than 50\% participants voted for it, or else ``0''. 

The stimuli containing reference and variant sentences belong to either of the orderings: \emph{Canonical} or \emph{Non-canonical}. 
Table \ref{tab:human-eval} presents the results of our experiment. Overall, of 167 human-validated pairs, 89.92\% of the reference sentences originally appearing in the HUTB corpus were also preferred by native speakers compared to the artificially generated variants expressing the very same proposition. Across all construction types, the full model was better at predicting human preferences (76.65\%) than it was at predicting the corpus reference sentences (74.85\%). Furthermore, Pearson's correlation between classifier predictions and human judgments was 0.534, and between classifier predictions and corpus labels was 0.497. 
Moreover, across all of the analyzed verb classes, the classifier using all measures was as good or better at predicting human choices than it was at discriminating reference from variant sentences, indicating a promising ability for these measures to reproduce human behavior. Further work is required to tease apart the relative contributions of the different predictors in modelling human choices.

\section{Discussion}\label{sect:disc}

Written text is a consequence of language production and is often edited to facilitate comprehension for the readers. According to Levelt's (\citeyear{levelt1989speaking}) language production model, speakers evaluate their own utterances by comprehending their own speech and make necessary adjustments to an utterance via a self-monitoring loop. Therefore, we interpret our results in the light of the \textsc{dual mechanism account}~\citep{tooley2010syntactic}  described earlier in the introduction. This account makes claims pertaining to both production and comprehension and~\citet{tooley2014parity} demonstrates the parity of syntactic persistence across both phenomena. Our results indicate that the dual mechanism account can be extended to postulate a viable model of priming effects in Hindi word order. Constituent ordering choices demonstrate both lexically independent syntactic priming as well as lexically dependent effects. We discuss how these two effects are induced by distinct underlying mechanisms (as stated at the outset), {\em viz.}, implicit learning ~\citep{BockGriff2000,chang2006}, and residual activation~\citep{Pickering1998} respectively.

Previous work suggests that lexical overlap between prime and target sentences enhances syntactic priming~\citep{Pickering1998,gries2005}. We also show that certain verb classes are more susceptible to priming than others. Specifically, \textsc{give} verbs selecting double objects are most prone to priming, a case demonstrated in English as well~\citep{gries2005}, thus providing cross-linguistic support for the finding. Hindi conjunct verbs in prime sentences trigger subsequent target sentences with conjunct verbs, and preverbal word order patterns for Hindi conjunct verbs are influenced by the repetition of lexical cues mentioned in the previous sentence. These two findings lend credence to the idea in the literature that lexical boost effects are attested for heads (conjunct verbs in this case) as well as other non-head lexical items~\citep{reitter2011}. The explanation for such effects stems from the residual activation theory~\citep{Pickering1998} where activated lemmas (linguistic category and combinatory nodes) in the prime utterance retain their activation for a short time. The residue of such activation is transferred to the target lemma.~\citet{reitter2011} proffer an alternative explanation for lexical boost via spreading activation mechanism posited by the ACT-R framework of cognition.

However, we observe syntactic priming independent of lexical effects over and above lexical repetition in double object constructions. Our verb-specific priming analyses indicate that prime sentences need not share the same main verb as the target sentence; instead, successive sentences may have a similar argument structure (subcategorization frame), which enforces a tendency to repeat canonical structures. \citet{tooley2010syntactic} show that such effects are best explained by the implicit learning account~\citep{BockGriff2000,chang2006}, where language users unconsciously acquire abstract routines over a period of time. In stark contrast to short-lived residual activation accounting for lexical boost effects,~\citet{BockGriff2000} showed that lexically independent syntactic priming effects persisted even when 10 intervening structures occurred between prime and target utterances. The relationship between prediction (quantified using our surprisal measures) and learning is made explicit in the P-chain framework of~\citet{dell-garden-path} connecting production and comprehension. According to P-chain assumptions, prediction error leads to implicit learning, which in turn helps the prediction system to adapt to less common structures (like double object constructions), which are known to induce higher priming strengths compared to commonplace structures~\citep{Ferreira2003b,JaegerSnider2007,bernolet2010}.

While our results demonstrate priming at the level of verb classes,~\citet{husain-yadav2020} showed that the combinatory properties of the verb need not be the sole driver of priming in Hindi. In their self-paced reading experiments involving identical critical verbs in both prime and target sentences, they observed faster reading times only in the target condition where nominals were marked by a locative case marker (in contrast to accusative and ergative conditions). Language-specific properties like case markers and the relationship between Hindi production and comprehension processes needs to be investigated more thoroughly by extending our preliminary human evaluation (via a simple forced choice task) using more fine-grained measures like reading aloud and silent reading times as proposed by~\citet{scil:sid}. 

Overall, in line with the assumptions of the \textsc{dual mechanism account}, our main findings suggest that Hindi word order choices are influenced by both lexically independent syntactic priming effects as well as lexically dependent priming effects. 
Future inquiries need to explore controlled experiments to corroborate the psychological reality of our current results.

\section*{Acknowledgements}

We would like to thank the first author's dissertation committee members: Dr Mausam and Dr Samar Husain, and the members of the Cornell's C.Psyd research group for their invaluable comments and feedback on this work. We thank Cornell's ethics board for their approval on human data collection for this project and Rupesh Pandey for his logistical help in collecting human judgement data associated with this work. 
We are also indebted to the anonymous reviewers of AACL 2022, ACL ARR 2021 and COMCO 2021 for their detailed and insightful feedback. Finally, the last two authors acknowledge extramural funding from the Cognitive Science Research Initiative, Department of Science and Technology, Government of India (grant no. DST/CSRI/2018/263). 

\bibliographystyle{aclnatbib}
\bibliography{bibfile,extra}

\pagebreak

\appendix

\section*{Appendix}

\section{Variant Generation}\label{appendix:A}

\begin{figure*}[t]
\centering
\noindent\makebox[\textwidth]{%
\subfloat[Figure][Dependency tree]{
\label{fig:intro-tree}
\includegraphics[scale=0.55]{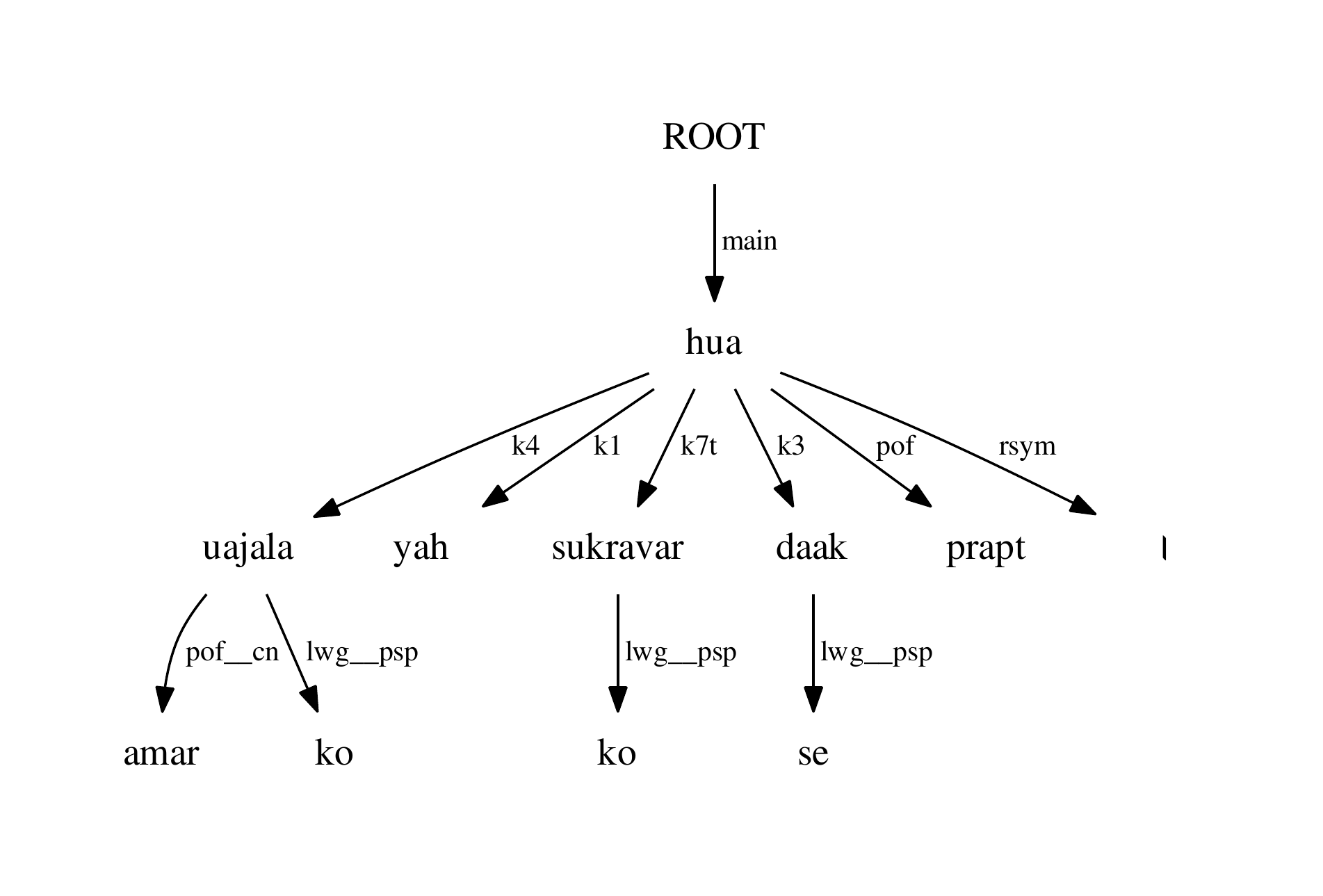}
}
\subfloat[Table][Dependency relations]{
\label{tab:intro-tree-labels}
  \begin{scriptsize}
\begin{tabular}[b]{lc}
\hline 
Label & Dependency \\
      & relation\\\hline
{\it Invariant syntactic relations}&\\
k1 & subject/agent\\
k2 & object/patient\\
k3 & instrument\\
k4 & object/recipient\\
k7t & location in time\\\hline
{\it Complex predicate relation}&\\
pof & parts of\\
          & conjunct verb\\
pof\_cn &  parts of\\
          & compound noun\\\hline
{\it Local word group (lwg)} &\\
lwg\_psp & postposition\\
lwg\_vaux & auxilliary verb\\\hline
{\it Symbols} &\\
rsym & symbol relation\\
\end{tabular}
   \end{scriptsize}
}}
\caption{Example HUTB dependency tree and relation labels}
\label{fig:hutb-tree}
\end{figure*}

\begin{exe}

  \ex 
  {\label{ex:hindi-prev-sent-app} \textbf{Context sentence}
    \gll {\underline{amar ujala}-ki} {bhumika} {nispaksh} {rehti} {hai}\\
         {Amar Ujala-\textsc{gen}} {role} {unbiased} {remain} {be\textsc{.prs.sg}}\\
     \glt Amar Ujala's role remains unbiased.}  

\end{exe}

\begin{exe}

  \ex \label{ex:hindi-intro-app}
  \begin{xlist}
  \ex[]{\label{ex:hindi-order-ref-app}
    \gll {\underline{amar ujala}-ko} {\bf yah} {\emph{sukravar}-ko} {daak-se} {prapt} {hua} [Given-Given = 0] \textbf{(Reference)}\\
         {\underline{Amar Ujala}-\textsc{acc}} {\bf it} {\emph{friday}-on} {post-\textsc{inst}} {receive} {be\textsc{.pst.sg}}\\
     \glt \underline{Amar Ujala} received \textbf{it} by post on \emph{Friday}.}
  \ex[] {\label{ex:hindi-order-var1-app} {{\bf yah} \underline{{amar ujala}}-ko \emph{sukravar}-ko daak-se prapt hua} [Given-Given = 0] \textbf{(Variant 1)}}
  \ex[] {\label{ex:hindi-order-var2-app} {\emph{sukravar}-ko {\bf yah} \underline{amar ujala}-ko daak-se prapt hua} [New-Given = -1] \textbf{(Variant 2)}}
  \end{xlist}

\end{exe}

This work uses sentences from the Hindi-Urdu Treebank (HUTB) corpus of dependency trees~\citep{Bhatt2009} containing well-defined subject and object constituents. Figure \ref{fig:hutb-tree} displays the dependency tree (and a glossary of relation labels) for reference sentence \ref{ex:hindi-order-ref-app}. The grammatical variants were created using an algorithm that took as input the dependency tree corresponding to each HUTB reference sentence. The re-ordering algorithm permuted the preverbal dependents of the root verb and linearized the resulting tree to obtain variant sentences. For example, corresponding to the reference sentence \ref{ex:hindi-order-ref-app} and its root verb ``hai'' (see figure \ref{fig:intro-tree}), the preverbal constituents\footnote{Hindi is not a strictly verb-final language but the majority of the constituents in the HUTB corpus are preverbal. \citet{cog:sid} in their corpus analysis with 13274 HUTB sentences found 20,750 pairs of preverbal constituents and 2599 pairs of postverbal constituents. Therefore, we also limit our variant generation (via reordering of constituents) and subsequent experiments on word-order variation in the preverbal domain only and leave the postverbal constituents in the reference-variants sentences as it is.} with parents as ``ujala'', ``yah'', ``suravar'', ``daak'', and ``prapt'' were permuted to generate the artificial variants (\ref{ex:hindi-order-var1-app} and \ref{ex:hindi-order-var2-app}). The ungrammatical variants were automatically filtered out using dependency relation sequences (denoting grammar rules) attested in the gold standard corpus of HUTB trees. In the dependency tree \ref{fig:intro-tree}, ``k4-k1'', ``k7t-k1'', ``k3-k7t'', and ``pof-k3'' are dependency relation sequences. 
In cases where the total number of variants exceeded 100 (a random cutoff),\footnote{Higher and lower cutoffs do not affect our results.} we chose 99 non-reference variants randomly along with the reference sentence.

\section{Information Status Annotation}\label{appendix:B}

The subject and object constituents in a sentence were assigned a \textit{Given} tag if any content word within them was mentioned in the preceding sentence or if the head of the phrase was a pronoun. All other phrases were tagged as \textit{New}. The sentence example \ref{ex:hindi-intro-app} illustrates the proposed annotation scheme. 

\begin{itemize}

\item Example \ref{ex:hindi-order-ref-app} follows {\it Given-Given} ordering --- The object ``Amar Ujala'' in the sentence is mentioned in the preceding context sentence \ref{ex:hindi-prev-sent-app}, it would be annotated as \emph{Given}. In contrast, the subject ``yah'' is a pronoun so it would also be tagged as \emph{Given} following the annotation scheme. 

\item Example \ref{ex:hindi-order-var2-app} follows {\it New-Given} ordering --- The object ``sukravar" in the sentence should be tagged as \emph{New} as it is not mentioned in the preceding context sentence \ref{ex:hindi-prev-sent-app}. In contrast, the subsequent pronoun ``yah'' which acts as the subject of the sentence should be tagged as \emph{Given} following the annotation scheme. 

\end{itemize}

\section{Adaptation Learning Rate}\label{appendix:lr-adapt}

Table~\ref{tab:lr-rate} illustrates the results of our learning rate experiments. Interestingly, \protect\citet{van2018neural} found that an adaptive learning rate of 2 minimized validation perplexity in English as well, though we leave further investigation of this to future work.

\begin{table*}[ht]
\centering
\scalebox{1.1}{
\begin{tabular}{|l|c|c|c|c|c|c|c|}
\hline
\textbf{Learning Rate} & 0 & 0.002 & 0.02 & 0.2 & \textbf{2} & 20 & 200 \\ \hline
\textbf{Perplexity} & 103.29 & 98.79 & 87.78 & 66.64 & \textbf{56.86} & 117.91 & $\sim10^9$ \\ \hline 
\end{tabular}}
\caption{Learning rate influence on lexical and syntactic adaptation for the validation set containing 13274 sentences (the initial non-adaptive model performance is when we use a learning rate of 0)}
\label{tab:lr-rate}
\end{table*}

\section{Correlation Plot}\label{appendix:pearson}

\begin{figure*}[!htbp]
    \centering
    \scalebox{0.61}{
    \includegraphics{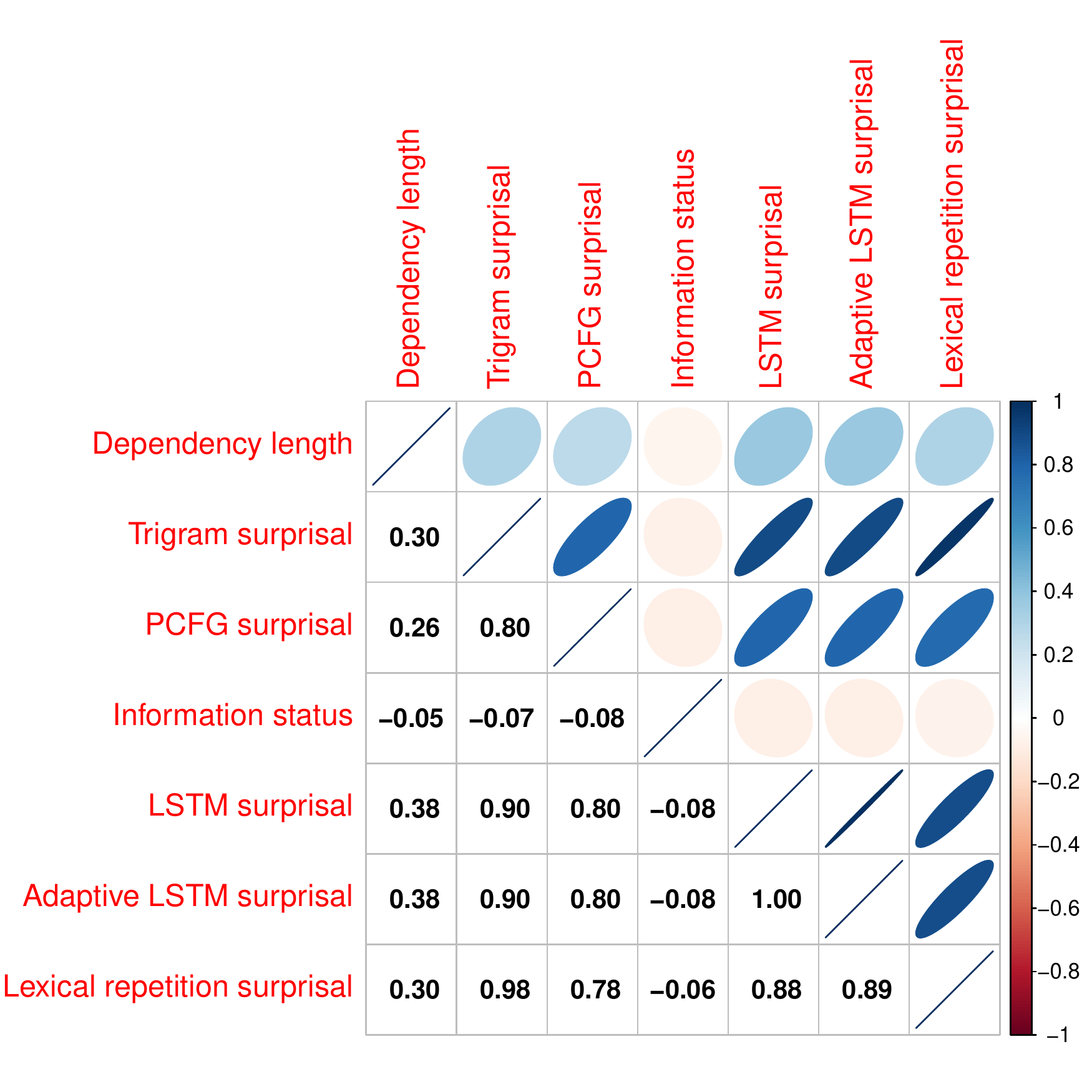}}
    \caption{Pearson’s coefficient
of correlation between different pairs of predictors}
    \label{fig:corr-plot}
\end{figure*}

The Pearson's correlation coefficients between different predictors are displayed in Figure \ref{fig:corr-plot}. The adaptive LSTM surprisal has a high correlation with all other surprisal features and a low correlation with dependency length and information status score.

\section{\textsc{give} Verb Class Regression Model}\label{appendix:verb-reg}

Table \ref{tab:regr-result-give} displays the results of our regression experiment predicting word order choices in Hindi on subset of the dataset involving Levin's verb class with lemma \textsc{give}.

\begin{table}[ht]
\centering
\scalebox{0.91}{
\begin{tabular}{lccc}
\toprule
Predictor & $\hat\beta$ & $\hat\sigma$ & t\tabularnewline
\midrule
intercept  & \textbf{1.50} & 0.002 & 638.32 \\
trigram surprisal  & \textbf{-0.11} & 0.013 & -8.57 \\
dependency length  & \textbf{0.01} & 0.003 & 2.78 \\
pcfg surprisal  & \textbf{-0.08} & 0.004 & -18.87 \\
IS score  & \textbf{0.02} & 0.002 & 10.01 \\
lex-rept surprisal  & 0.01 & 0.012 & 0.46 \\
lstm surprisal  & \textbf{0.08} & 0.036 & 2.25 \\
adaptive lstm surprisal  & \textbf{-0.36} & 0.037 & -9.86\\
\bottomrule 
\end{tabular}}
\caption{Regression model on lemma verb \textsc{give} data set (14094 data points; all significant predictors denoted by $|$t$|$\textgreater{}2)}
\label{tab:regr-result-give}
\end{table}

\section{Levin's Verb Class and Case Density}\label{appendix:verbcd}

Table \ref{tab:verb:case} presents the case density and frequency distributions of different \citeauthor{levin1993english}'s verb classes in our dataset.

\begin{table}[ht]
\centering
\scalebox{0.85}{
\begin{tabular}{l|ccc}
\textbf{Verb Types} & \textbf{Case density} & \textbf{Freq} & \textbf{Freq (\%)} \\\hline
\textsc{give} & 0.45 & 372 & 18.64 \\
\textsc{do} & 0.39 & 726 & 36.37 \\
\textsc{communication} & 0.67 & 264 & 13.23 \\
\textsc{motion} & 0.39 & 93 & 4.66 \\
\textsc{social} & 0.4 & 242 & 12.12 \\
\textsc{perception} & 0.32 & 36 & 1.8 \\
\textsc{destroy} & 0.63 & 34 & 1.7 \\
\textsc{lodge} & 0.32 & 95 & 4.76 \\
\textsc{put} & 0.4 & 52 & 2.61 \\
\textsc{others} & 0.43 & 82 & 4.11 \\\hline
\textbf{Full} & 0.44 & 1996 & 100
\end{tabular}}
\caption{Levin's verb semantic classes and case density (i.e., number of case markers per constituent in a sentence)}
\label{tab:verb:case}
\end{table}

\section{Argument Ordering and Case Density}\label{appendix:k1k2k4cd}

\begin{table}[H]
\centering
\scalebox{0.92}{
\begin{tabular}{l|ccc}
\textbf{Alternation} & \textbf{Case density} & \textbf{Freq} & \textbf{Freq (\%)} \\\hline
{S-IO-DO} & 0.48 & 185 & 9.27 \\
{S-DO} & 0.39 & 1417 & 70.99 \\
{S-IO} & 0.59 & 394 & 19.74 \\\hline
\textbf{Full} & 0.44 & 1996 & 100 \\
\end{tabular}}

\caption{Argument ordering and case density (i.e., number of case markers per constituent in a sentence)}
\label{tab:arg:case}

\end{table}

\section{Levin's classes of verbs within Double Object (S-IO-DO) alternation}\label{appendix:verbclass}

Table \ref{tab:verb:k1k2k4} presents \citeauthor{levin1993english}'s syntactico-semantic classes of verbs within S-IO-DO data points from Table \ref{tab:verb-arg:pred}.

\begin{table*}[ht]
\centering
\scalebox{1}{
\begin{tabular}{|l|c|c|c|c|}
\hline
\textbf{Verb Lemma} & \textbf{Frequency} & \textbf{Freq (\%)} & \textbf{Verb Types} & \textbf{Freq (\%)} \\ \hline
\textit{chah} & 127 & 1.37 & \multirow{4}{*}{\textsc{social}} & \multirow{4}{*}{2.59} \\ \cline{1-3}
\textit{nawaja} & 5 & 0.05 &  &  \\ \cline{1-3}
\textit{mil} & 5 & 0.05 &  &  \\ \cline{1-3}
\textit{bech} & 104 & 1.12 &  &  \\ \hline
\textit{daal} & 99 & 1.07 & \multirow{3}{*}{\textsc{put}} & \multirow{3}{*}{2.13} \\ \cline{1-3}
\textit{jutaa} & 75 & 0.81 &  &  \\ \cline{1-3}
\textit{pilaa} & 23 & 0.25 &  &  \\ \hline
\textit{dikha} & 28 & 0.3 & \textsc{perception} & 0.3 \\ \hline
\textit{badal} & 99 & 1.07 & \textsc{lodge} & 1.07 \\ \hline
\textbf{de} & 3240 & 34.92 & \multirow{5}{*}{\textsc{\textbf{give}}} & \multirow{5}{*}{\textbf{57.82}} \\ \cline{1-3}
\textit{saup} & 1090 & 11.75 &  &  \\ \cline{1-3}
\textit{bhej} & 569 & 6.13 &  &  \\ \cline{1-3}
\textit{maang} & 419 & 4.52 &  &  \\ \cline{1-3}
\textit{dilaa} & 46 & 0.5 &  &  \\ \hline
\textit{kar} & 1737 & 18.72 & \multirow{4}{*}{\textsc{do}} & \multirow{4}{*}{24.03} \\ \cline{1-3}
\textit{karaa} & 465 & 5.01 &  &  \\ \cline{1-3}
\textit{chipaa} & 23 & 0.25 &  &  \\ \cline{1-3}
\textbf{ban} & 5 & 0.05 &  &  \\ \hline
\textit{kah} & 883 & 9.52 & \multirow{4}{*}{\textsc{communication}} & \multirow{4}{*}{12.06} \\ \cline{1-3}
\textit{sunaa} & 198 & 2.13 &  &  \\ \cline{1-3}
\textit{likh} & 23 & 0.25 &  &  \\ \cline{1-3}
\textit{bataa} & 15 & 0.16 &  &  \\ \hline\hline
\textbf{Full (S-IO-DO)} & 9278 & 100 & & 12.74\% of 72388\\ \hline
\end{tabular}}
\caption{\citeauthor{levin1993english}'s syntactico-semantic classes of verbs within S-IO-DO data points from Table \ref{tab:verb-arg:pred}}
\label{tab:verb:k1k2k4}
\end{table*}

\section{Information Profile: Syntactic Priming}\label{appendix:k1k2k4}

\begin{figure*}[ht]
    \begin{center}
    \includegraphics[width=1\textwidth]{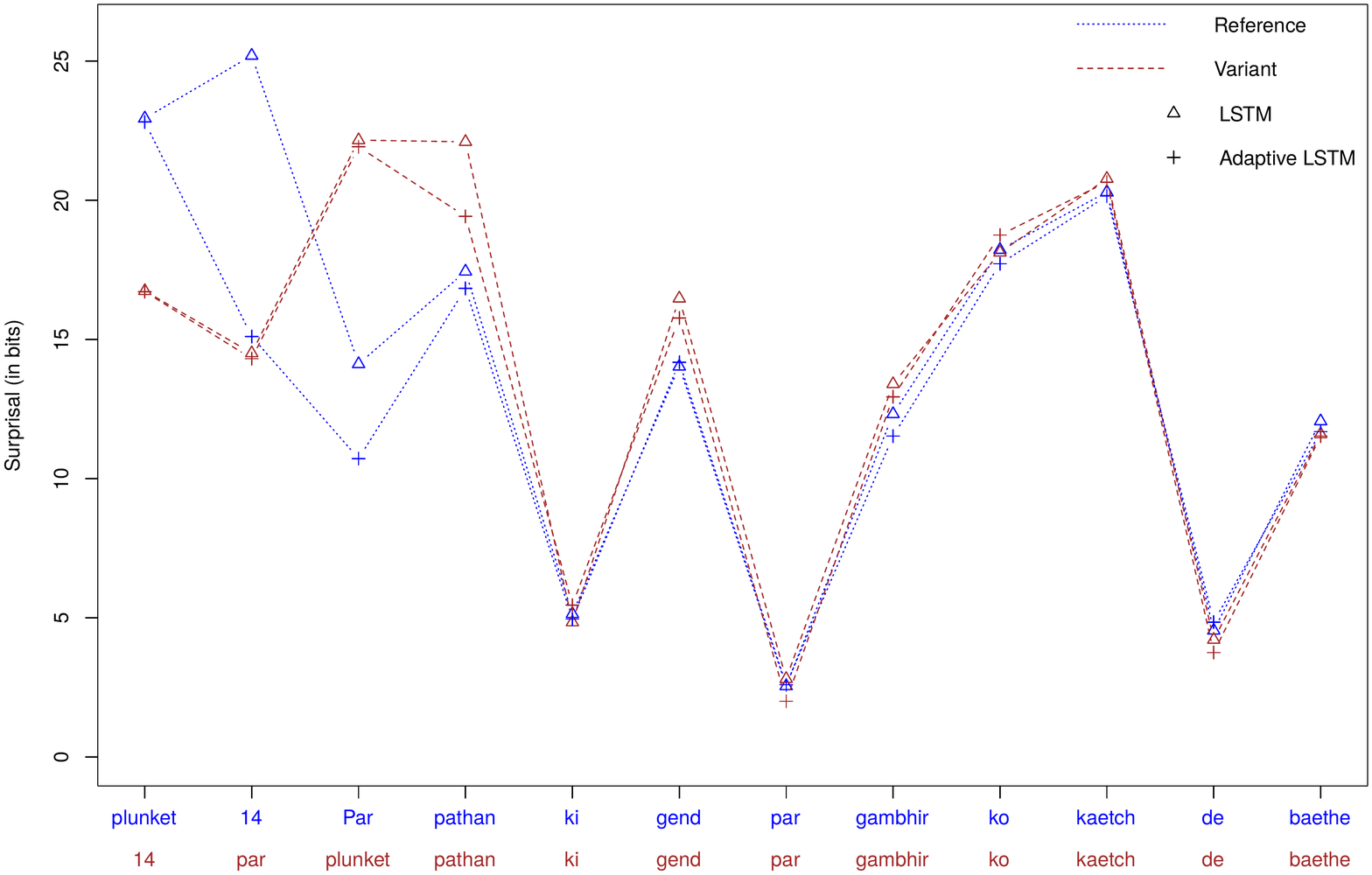}
    \end{center}
    \caption{Information profiles for the reference-variant pair \ref{ex:ref} and \ref{ex:var}}
    \label{fig:k1k2k4-profile}
\end{figure*}

Figure \ref{fig:k1k2k4-profile} displays the information profiles for the reference-variant pair \ref{ex:ref} and \ref{ex:var}.

\begin{table*}[ht]
\scalebox{0.77}{
\begin{tabular}{lcccccccc}
\multicolumn{2}{c}{Type} & {Trigram surp} & Deplen & PCFG surp & IS score & LSTM surp & Adaptive LSTM surp & Lex rept surp \\\hline
Example \ref{ex:ref} & Reference & 34.27 & 24 & 107.04 & 0 & 173.06 & \textbf{156.88} & 36.45 \\
Example \ref{ex:var} & Variant & 33.92 & 23 & 105.11 & 0 & 171.49 & 165.86 & 36.45\\\hline
Example \ref{ex:ref-lex-rept} & Reference & 58.04 & 40 & 144.98 & -1 & 186.10 & 185.75 & \textbf{54.43}\\
Example \ref{ex:var-lex-rept} & Variant & 57.68 & 26 & 143.06 & 1 & 185.31 & 184.52 & 56.84 \\\hline
\end{tabular}}
\caption{Predictor scores for reference-variant pairs}
\label{tab:sent-level-scores}
\end{table*}

\section{Broad Coverage Analysis}\label{sect:full-data}

Our regression results over the entire data set (Table \ref{tab:regr-results1}) indicate that all the measures considered in our work are significant predictors of syntactic choice (\emph{i.e.,} classifying reference and variant sentences). The negative regression coefficients for all surprisal metrics indicate that log-odds of predicting the reference sentences increase with decrease in their surprisal values. In other words, corpus reference sentences have consistently lower surprisal scores compared with the artificially generated competing variants. And adding adaptive LSTM surprisal into a model containing all other predictors significantly improved the fit of our regression model ($\chi^2$ = 66.81; p $<$ 0.001). The positive regression coefficient for information status (IS) score indicates that reference sentences adhere to \textit{given-new} ordering. Similarly, adding IS score into a model containing all other predictors significantly improved the fit of our regression model ($\chi^2$ = 127.94; p $<$ 0.001). However, the positive regression coefficient of dependency length suggests that reference sentences exhibit \emph{longer} dependency lengths compared to their variant counterparts, violating locality considerations. This further conjectures that dependency length might be in conflict with (and/or overridden by) other factors like discourse and priming. Future work needs to investigate if word-order preferences can be jointly optimized using multiple factors~\citep{GildeaJaeger2015}.

We now examine the relative performance of each predictor in classifying reference sentences against the paired counterfactual grammatical variant by estimating the prediction accuracy (i.e., the percentage of data points where the model chose the reference sentence as the best choice compared to the paired variant). We performed 10-fold cross-validation, trained the model on 9 folds, and generated its prediction on the remaining fold. Table \ref{tab:lex-adapt-pred-acc} presents the individual as well as collective prediction performance of our predictors. Among individual predictor performances (Left side of Table \ref{tab:lex-adapt-pred-acc}; Full data), both adaptive and non-adapt LSTM surprisal achieved the highest classification accuracy. However, over a baseline model comprising every other predictor, adaptive LSTM surprisal induced a significant boost of 0.03\% in classification accuracy (p = 0.04 using McNemar's two-tailed test) only when lexical repetition surprisal was not included in the model. 

\begin{table*}
\centering
\scalebox{1}{
\begin{tabular}{lccc}
\toprule
Predictor & $\hat\beta$ & $\hat\sigma$ & t\tabularnewline
\midrule 
intercept  & 1.50  & 0.001  & 1496.47 \\
trigram surprisal  & -0.08  & 0.005  & -14.53 \\
dependency length  & 0.02  & 0.001  & 15.55 \\
pcfg surprisal  & -0.07  & 0.002  & -39.46 \\
IS score  & 0.01  & 0.001  & 11.32 \\
lex-rept surprisal  & -0.03  & 0.005  & -5.31 \\
lstm surprisal  & -0.14  & 0.016  & -9.26 \\
adaptive lstm surprisal  & -0.13  & 0.016  & -8.18\\
\bottomrule 
\end{tabular}}
\caption{Regression model on full data set ($N=72833$; all significant predictors denoted by $|$t$|$\textgreater{}2)}
\label{tab:regr-results1}
\end{table*}

\begin{table*}
\centering
\scalebox{0.8}{
\begin{tabular}{|l|c|c|c|c|c|c|}
\hline
\textbf{Predictors} & \textbf{\begin{tabular}[c]{@{}c@{}}Full\\ Accuracy \%\end{tabular}} & \textbf{Conjunct Verb} & \multirow{8}{*}{} & \textbf{Predictors} & \textbf{\begin{tabular}[c]{@{}c@{}}Full\\ Accuracy \%\end{tabular}} & \textbf{Conjunct Verb} \\ \cline{1-3} \cline{5-7} 
a = IS score & 51.84 & 52.08 & &  \multicolumn{3}{c|}{Collective: with repetition effects} \\ \cline{1-3} \cline{5-7} 
b = dep length & 62.31*** & 66.32*** &  & base1 = a+b+c+d+e+f & \textbf{95.05} & \textbf{96.33} \\ \cline{1-3} \cline{5-7} 
c = pcfg surp & 86.86*** & 89.20*** &  & base1 + g & 95.06 & 96.34 \\ \cline{1-3} \cline{5-7} 
d = lex repetition surp & 90.07*** & 92.69*** & & \multicolumn{3}{c|}{\multirow{2}{*}{Collective: without repetition effects}} \\ \cline{1-3}
e = 3-gram surp & 91.18*** & 93.54*** & &  \multicolumn{3}{c|}{} \\ \cline{1-3} \cline{5-7} 
f = lstm surp & \textbf{94.01***} & \textbf{95.67***} &  & base2 = a+b+c+e+f & 95.06 & 96.34 \\ \cline{1-3} \cline{5-7} 
g = adaptive lstm surp & 94.06 & 95.68 &  & base2 + g & \textbf{95.09*} & \textbf{96.38*} \\ \hline
\end{tabular}}
\caption{Prediction performances (Full data set (72833 points), Conjunct Verb (51617 points); each row refers to a distinct model; *** McNemar's two-tailed significance compared to model on previous row)}
\label{tab:lex-adapt-pred-acc}
\end{table*}


\end{document}